\PassOptionsToPackage{table}{xcolor} 
\documentclass[cameraready]{Interspeech}

\usepackage{arydshln}
\usepackage{xcolor}
\usepackage{multirow}
\usepackage{subcaption}

\title{Search-GRT: \underline{G}uided \underline{R}etrieval \underline{T}raining of \underline{Search} Agents to Optimize for Complex Question Answering}

\author[correspondingauthor]{Aounon}{Kumar}
\author[]{Sudipta}{Paul}
\author[]{Vivek}{Kulkarni}
\author[]{Vijay}{Srinivasan}
\author[]{Srinivas}{Chappidi}

\address{AI Center-Mountain View, Samsung Electronics}

\email{\{aounon.kumar, sudipta.paul, v.kulkarni1, v.srinivasan, vasu.c\}@samsung.com}


\usepackage{comment}

\newcommand{\SizePerGT}{\kappa}

\begin{document}

\maketitle

\begin{abstract}
The effective use of search engines by large language models (LLMs) remains a significant challenge, particularly in complex, multi-hop question-answering (MHQA) tasks. These tasks require the model to decompose questions into subqueries, retrieve relevant information, and synthesize answers from multiple sources, often leading to cascading errors due to poor retrieval in early stages. Reinforcement learning (RL) has shown promise in improving LLMs' search capabilities, but it often suffers from sparse rewards during training, hindering the model's ability to learn effectively.
To address these challenges, we introduce Guided Retrieval Training (GRT), a novel method that improves the performance of a search agent by restricting the retrieval process during RL training using ground truth information. By focusing on a curated set of relevant documents, GRT provides the model with a stronger learning signal, mitigating the problem of sparse rewards and improving its ability to generate accurate subqueries and synthesize correct answers. Our experimental results demonstrate that GRT achieves consistent performance improvements over existing methods, such as Search-R1, across a wide range of question-answering (QA) tasks. Notably, GRT excels in MHQA tasks, achieving over 40\% improvements in performance. Additionally, GRT enhances training efficiency by achieving better QA performance with fewer training steps.
\end{abstract}

\section{Introduction}

The advent of large language models (LLMs) has revolutionized the way we search for and interact with information on the internet. Traditionally, users were required to manually compile information by reviewing a list of results generated by search engines.
The integration of LLMs into search systems has significantly transformed this landscape, enabling more intuitive and efficient information retrieval.
Initially, LLMs were employed to enhance search by rewriting user queries to improve search results \cite{MaGHZD23,ma-etal-2023-query,ye-etal-2023-enhancing,liu2025genrewrite}.
Subsequently, LLMs were used to summarize and compile search results into coherent responses, directly addressing users' specific inquiries \cite{Fan2024RAGSurvey,Gao2023RAGsurvey,search_arena_2025}. This development marked a significant step toward more user-friendly and effective search systems.

The next step in this evolution was the emergence of search agents.
These are LLMs capable of using search engines to answer complex user queries. They are designed to dynamically query the search engine multiple times, retrieve relevant information, and synthesize a final answer from the compiled results.
Despite their potential, search agents face significant challenges in effectively utilizing search engines to retrieve and synthesize information for accurate answers. 
Reinforcement learning (RL) has emerged as a promising approach to enhance LLMs' ability to use search engines effectively \cite{jin2025searchr}.
Unlike naive prompting-based methods, RL trains LLMs to use search engines more effectively, enabling them to retrieve relevant information and synthesize coherent answers from the retrieved information.
However, RL training can suffer from a poor reward signal during the initial stages of training.
The untrained LLM often produces suboptimal queries, leading to the retrieval of irrelevant or incomplete information, which in turn results in incorrect answers.
This problem is particularly pronounced in complex question-answering tasks, such as multi-hop question-answering (MHQA), where the model must decompose the input query into multiple subqueries, retrieve relevant information for each subquery, and integrate it to synthesize the final answer.

Poor retrieval in the initial stages can propagate errors to subsequent query formulation, creating a cascading effect that degrades the overall performance of the system.
For instance, if the LLM fails to retrieve accurate information for an early subquery, it may generate incorrect subsequent subqueries, further compounding the problem. As a result, the synthesized final answer is often incorrect, and the LLM receives minimal or no reward during RL training. This lack of meaningful learning signals hinders the model's ability to improve its query formulation and reasoning capabilities.



To address these challenges, we introduce Guided Retrieval Training (GRT), a search agent training method that improves  performance by restricting the retrieval process during training using ground truth information. By focusing on a curated set of relevant documents, GRT provides the model with a stronger learning signal, mitigating the problem of sparse rewards and enhancing its ability to generate accurate subqueries and synthesize correct answers.
Our main contributions include:
\begin{enumerate}
\item A novel method that enhances search agents by restricting retrieval during training using ground truth information.
\item Experimental results demonstrating consistent performance improvements over baselines across several QA tasks, with significant gains in multi-hop QA.
\item Further analysis showing GRT's ability to improve retrieval and answer accuracy, while generalizing to unseen samples and utilizing the actual search engine at inference time.
\end{enumerate}


The implications of our work extend beyond search agent research, contributing to the broader goal of improving AI systems' ability to retrieve and synthesize information from vast knowledge bases. By introducing a guided training mechanism, GRT sets a new benchmark for search agent research and paves the way for further innovations in AI training strategies.

\section{Related Work}
Recent research has leveraged LLMs as search-augmented agents, enabling open-domain question answering by retrieving documents from large corpora and generating answers based on the retrieved information \cite{chen2017reading}. Question answering tasks can be categorized into two sub-classes: general question answering \cite{kwiatkowski2019natural} and multi-hop question answering \cite{yang2018hotpotqa}. To address open-domain question answering, several approaches have been proposed such as IRCoT \cite{trivedi2023interleaving}, ReAct \cite{yao2022react}, CoRAG \cite{wang2025chain} and DeepRAG \cite{guan2025deeprag}. Recently, Reinforcement learning has been used to optimize end-to-end agent trajectories directly. Search-R1 \cite{jin2025searchr} shows that purely outcome-based RL can train LLMs to interleave reasoning with multiple search queries. Other notable works include HiPRAG \cite{wu2025hiprag}, R1-Searcher \cite{song2025r1}, $\beta$-GRPO \cite{wu2025search} and ZeroSearch \cite{sun2025zerosearch}.

\section{Our Method: Guided Retrieval Training}
\label{sec:grt}

During RL training, a search agent starts with an example user query from the training set. It breaks the query into subqueries and uses the search engine to find relevant information for each subquery.
The search engine searches over a large collection of documents—in our case, all Wikipedia articles up to 2018—to find documents related to the search query. The search agent reviews the retrieved information, creates more subqueries if needed, and finally combines all the information to form the final answer.
If the final answer matches one of the correct answers (ground truth answers), the agent gets a positive reward. If not, it gets no reward.
However, in the early stages of RL training, the untrained agent struggles to create good subqueries. This leads to the retrieval of irrelevant or incomplete information, which disrupts the entire process and results in incorrect answers.
To solve this problem, our Guided Retrieval Training (GRT) method steps in during the search engine's retrieval process.
It restricts the search engine to a subset of highly relevant documents from the entire corpus.
This guides the process to return more accurate and relevant information.


Specifically, for a training example $s$, we obtain a set of ground truth information $\mathsf{GT}_s$ from the example and use it to restrict the search corpus to a small subset of documents that are the most similar to the ground truth information.
Mathematically, the restricted corpus $\mathsf{ResCorp}(s)$ is defined as
\[\mathsf{ResCorp}(s) = \bigcup_{i \in \mathsf{GT}_s} \{ d \in \Gamma \mid \text{similarity$(d, i) \in$ top-$\SizePerGT$ values}\},\]
where $\Gamma$ represents the entire corpus consisting of all Wikipedia documents till 2018 and $\SizePerGT$ is the number of documents included in the restricted corpus for each ground truth information $i \in \mathsf{GT}_s$ allowing duplicates.
We use cosine similarity of E5 text embeddings as the similarity metric.
By limiting the retrieval options during training using ground truth information, GRT helps the agent learn better. This way, it can generate better subqueries and generate correct answers more effectively.

We obtain the ground truth information needed for this purpose in the following two manners.
For the dataset HotpotQA, which provides ground truth passages in the training data, we directly utilize these passages to form the restricted corpus.
These passages represent documents that contain the necessary information to derive the ground truth answers.
For the dataset Natural Questions, which does not provide ground truth passages, we create the desired ground truth information by concatenating the user query and the ground truth answer for each ground truth answer in the sample.

\textbf{Text generation:} A search agent generates text through an interleaved sequence of reasoning, search, and retrieval steps~\cite{jin2025searchr}.
This structured approach ensures that the agent dynamically retrieves relevant information and integrates it into its reasoning process to produce accurate and contextually appropriate answers.
Given a user query, the agent begins by generating a sequence of reasoning tokens enclosed within \texttt{<think>} and \texttt{</think>} tags.
These tokens represent the agent's initial thoughts or hypotheses about how to approach the query.
Based on its inital reasoning, it formulates a subquery and places it inside \texttt{<search>} and \texttt{</search>} tags.
This triggers the search engine to retrieve documents relevant to the subquery.
The information fetched by the search engine is then placed inside \texttt{<information>} and \texttt{</information>} tags.
The agent iterates over the reasoning and retrieval steps multiple times, generating additional reasoning tokens, formulating new subqueries, and retrieving more relevant documents.
In the end, it synthesize all the retrieved information and produces the final answer between \texttt{<answer>} and \texttt{</answer>} tags.

\textbf{RL framework:} The search agent is trained using a reinforcement learning framework, where the LLM powering the search agent is used as the policy model.
The training objective is to maximize the following function:
\begin{equation*}
\max_{\theta}
\mathbb{E}_{x \sim \mathcal{D},y \sim \pi_\theta(\cdot \mid x; R)}
\left[r(x,y)\right]
- \beta \mathbb{D}_{\mathrm{KL}}\left(\pi_\theta \parallel \pi_{\mathrm{ref}}\right),
\end{equation*}
where $\pi_\theta$ represents the policy model with optimizable parameters $\theta$, $x$ is a user query drawn from a distribution $\mathcal{D}$ and $y$ is the sequence of tokens generated by the policy model, conditioned on the input $x$ and the information retrieved by the search engine $R$.
$r(x,y)$ represents the reward received by the policy model for a given input $x$ and a generated sequence $y$.
Following the setup in~\cite{jin2025searchr}, we set the reward to be 1 when the generated answer exactly matches one of the ground truth answers and 0 otherwise.
$\mathbb{D}_{\mathrm{KL}}\left(\pi_\theta \parallel \pi_{\mathrm{ref}}\right)$ represents the Kullback-Leibler (KL) divergence of the policy model's distribution $\pi_\theta$ from that of a reference model $\pi_{\mathrm{ref}}$, which is used to ensure the policy model does not diverge significantly during training.
$\beta$ is a hyperparameter that controls the trade-off between reward maximization and policy regularization.
The reference model is fixed to a prior state of the policy model earlier in the training process.

\begin{table*}[t!]
    \centering
    \scriptsize
    \setlength{\tabcolsep}{4pt}
    \renewcommand{\arraystretch}{1.2}
    \caption{Main results. The highest performance is set in bold face and the second highest is underlined. $^\dagger/^\star$ represents in-domain/out-domain datasets.}
    \label{tab:main}
    \begin{tabular}{lccc|cccc|cc}
        \toprule
        \textbf{Methods} & \multicolumn{3}{c}{\textbf{General QA}} & \multicolumn{4}{c}{\textbf{Multi-Hop QA}} & \multicolumn{2}{c}{\textbf{Average}} \\
        \cmidrule{2-10}
        & \textbf{NQ$^\dagger$} & \textbf{TriviaQA$^\star$} & \textbf{PopQA$^\star$} & \textbf{HotpotQA$^\dagger$} & \textbf{2wiki$^\star$} & \textbf{Musique$^\star$} & \textbf{Bamboogle$^\star$} & \textbf{MHQA} & \textbf{All QA} \\
        \bottomrule 
        Direct Inference & 0.106 & 0.288 & 0.108 & 0.149 & 0.244 & 0.020 & 0.024 & 0.109 & 0.134 \\
        CoT & 0.023 & 0.032 & 0.005 & 0.021 & 0.021 & 0.002 & 0.000 & 0.011 & 0.015 \\
        IRCoT & 0.111 & 0.312 & 0.200 & 0.164 & 0.171 & 0.067 & 0.240 & 0.161 & 0.181 \\
        Search-o1 & 0.238 & 0.472 & 0.262 & 0.221 & 0.218 & 0.054 & \textbf{0.320} & 0.203 & 0.255 \\
        RAG & 0.348 & 0.544 & 0.387 & 0.255 & 0.226 & 0.047 & 0.080 & 0.152 & 0.270  \\
        SFT & 0.249 & 0.292 & 0.104 & 0.186 & 0.248 & 0.044 & 0.112 & 0.148 & 0.176  \\
        R1-base & 0.226 & 0.455 & 0.173 & 0.201 & 0.268 & 0.055 & 0.224 & 0.187 & 0.229  \\
        R1-instruct & 0.210 & 0.449 & 0.171 & 0.208 & \underline{0.275} & 0.060 & 0.192 & 0.184 & 0.224  \\
        Rejection Sampling & 0.294 & 0.488 & 0.332 & 0.240 & 0.233 & 0.059 & 0.210 & 0.186 & 0.265 \\
        Search-R1 & \underline{0.392} & \underline{0.575} & \textbf{0.440} & \underline{0.293} & 0.253 & \underline{0.082} & 0.194 & \underline{0.206} & \underline{0.318}  \\ 
        \rowcolor{gray!30} Search-GRT (Ours) & \textbf{0.435} & \textbf{0.589} & \underline{0.412} & \textbf{0.371} & \textbf{0.391} & \textbf{0.142} & \underline{0.282} & \textbf{0.297} & \textbf{0.375}  \\
        \bottomrule
    \end{tabular}
\end{table*}

\section{Experimental Setup}

\subsection{Datasets and Metrics}
Following the convention from previous work on search agents~\cite{jin2025searchr}, we use a collection of question answering (QA) datasets spanning two categories:
\begin{enumerate}
    \item \textbf{General Question Answering:} Natural Questions (NQ) \cite{kwiatkowski-etal-2019-natural}, TriviaQA \cite{joshi-etal-2017-triviaqa}, and PopQA \cite{mallen-etal-2023-trust}. These datasets contain examples of open-domain question answering tasks sourced from Google search queries, trivia games, and less popular factual knowledge.
    \item \textbf{Multi-Hop Question Answering (MHQA):} HotpotQA \cite{yang-etal-2018-hotpotqa}, 2WikiMultiHopQA (2Wiki for short) \cite{ho-etal-2020-constructing}, Musique \cite{trivedi-etal-2022-musique}, and Bamboogle \cite{press-etal-2023-measuring}. These datasets focus on complex question-answering tasks that require decomposing questions into subqueries and reasoning over multiple documents to synthesize the final answer.
\end{enumerate}
We use NQ and HotpotQA for training and use all datasets for evaluations.
Table \ref{tab:dataset_splits} shows the number of samples in the above datasets.

\begin{table}[t!]
    \centering
    \caption{Train, dev and test splits for datasets used in our experiments. For datasets that do not have a test split, the dev split is used for evaluations.}
    \label{tab:dataset_splits}
    \begin{tabular}{clrrr}
        \toprule
        \textbf{QA Category} & \textbf{Datasets} & \textbf{Train} & \textbf{Dev} & \textbf{Test} \\
        \midrule
        \multirow{3}*{General QA} & NQ	& 79168 & 8757 & 8757 \\
        & TriviaQA & 78785 & 8837 & 11313 \\
        & PopQA & N/A & N/A & 14267 \\
        \midrule
        \multirow{4}*{Multi-Hop QA} & HotpotQA & 90447 & 7405 & N/A \\
        & 2Wiki & 15000 & 12576 & N/A \\
        & Musique & 19938 & 2417 & N/A \\
        & Bamboogle & N/A & N/A & 125 \\
        \bottomrule
    \end{tabular}
\end{table}

Similar to the approach in \cite{jin2025searchr}, we use exact match (EM) as the correctness metric to evaluate the performance of our method.
The EM metric assesses whether the answer generated by the search agent exactly matches one of the ground truth answers in the dataset, using string matching for comparison.
Our evaluations include both in-ditribution (NQ and HotpotQA) and out-of distribution datasets (TriviaQA, PopQA, 2Wiki, Musique, Bamboogle) for each category of QA tasks.
This allows for a comprehensive assessment of our method across diverse question-answering tasks.
We report the performance for each individual task, as well as the average across all tasks.
Given that our approach is specifically designed to enhance multi-hop question-answering capabilities, we also report the average performance across all MHQA tasks.
This targeted analysis highlights the efficacy of our method in addressing the unique challenges associated with MHQA tasks.

\subsection{Baselines}
The primary baseline we compare against is Search-R1 \cite{jin2025searchr}. The performance numbers reported in Table~\ref{tab:main} for Search-R1 have been re-computed using the codebase provided in \cite{jin2025searchr} to ensure consistency and comparability.
We also include additional baselines from \cite{jin2025searchr}, categorized as follows:

\begin{enumerate}
\item \textbf{Methods without Retrieval:} Direct inference and Chain-of-Thought (CoT) reasoning \cite{cot_wei_2022}. These methods solely rely on the knowledge encoded in the LLMs weights to answer queries, without utilizing any external search engine.
They are inherently limited in tasks requiring information beyond the LLM's pre-trained knowledge, leading to suboptimal performance.
\item \textbf{Methods using Retrieval:} Retrieval-Augmented Generation (RAG) \cite{rag_lewis_2020}, IRCoT \cite{TrivediBKS23}, and Search-o1 \cite{Li-search-o1-2025}. These methods integrate a search engine with the LLM to retrieve relevant external information and augment the generation process. These methods improve the LLM's ability to answer complex queries by accessing up-to-date or domain-specific knowledge, improving accuracy and relevance.
\item \textbf{Fine-Tuning-based Methods:} Supervised fine-tuning (SFT) \cite{ChungHLZTFL00BW24}, RL-based fine-tuning without a search engine (R1) \cite{GuoYZSWZXZMBZY025} and rejection sampling \cite{yuan-rft-2023} with a search engine.
These methods improve the LLM's performance through fine-tuning on QA task data or the LLM's own responses that lead to correct answers.
\end{enumerate}
Our method (GRT) and Search-R1 fall at the intersection of the last two categories.

\subsection{Implementation Details}

To ensure a fair comparison, we adopt an experimental setup similar to that of \cite{jin2025searchr}. Our experiments are conducted using the 3 billion parameter base model from the Qwen-2.5 class of language models \cite{Yang-qwen-2024}. For RL training of the search agent, we use the Proximal Policy Optimization (PPO) algorithm \cite{ppo-SchulmanWDRK17} for 600 steps.
The search engine in our setup uses a dense retriever \cite{karpukhin-etal-2020-dense} with E5 text embeddings \cite{Wang-e5-2022}. The knowledge corpus used for the search engine retrieval mechanism consists of the 2018 Wikipedia documents.
For our restricted retrieval mechanism (see Section \ref{sec:grt}), we set the number of relevant documents per ground truth information $\SizePerGT$ to 300.

\section{Results and Analysis}
Our main results are presented in Table \ref{tab:main}, which compares the performance, as measured by the exact match metric, of our method with several baselines, including Search-R1.
Our method demonstrates strong performance across all individual QA tasks, achieving the highest or second-highest accuracy in each case. Furthermore, it outperforms all baselines in terms of the average performance across all tasks, highlighting its overall effectiveness.
Notably, for multi-hop question-answering (MHQA) tasks, our approach achieves a significant improvement of over 40\% compared to Search-R1, which is the second-best method. This substantial gain underscores the efficacy of our method in addressing the unique challenges of MHQA, such as decomposing complex questions into subqueries and synthesizing information from multiple retrieved documents.


\begin{figure}[t!]
    \centering
    \begin{subfigure}[t]{0.23\textwidth}
        \centering
        \includegraphics[width=\textwidth]{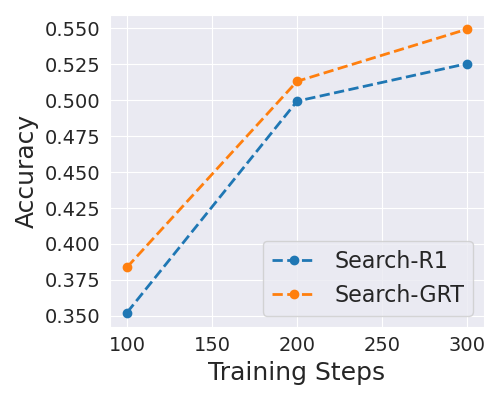}
        \caption{Retrieval Accuracy}
        \label{fig:ret_acc}
    \end{subfigure}%
    ~ 
    \begin{subfigure}[t]{0.23\textwidth}
        \centering
        \includegraphics[width=\textwidth]{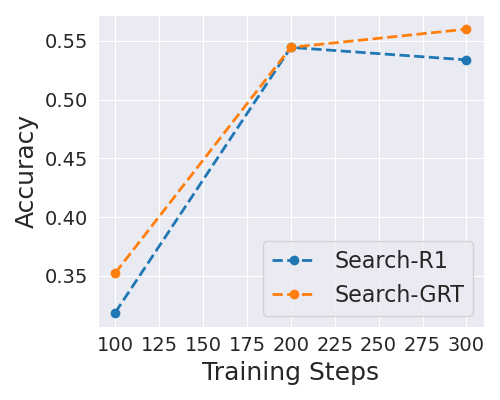}
        \caption{Answer Accuracy given Correct Retrieval}
        \label{fig:ans_acc}
    \end{subfigure}
    \caption{Comparing retrieval accuracy and answer correctness given correct retrieval of Search-R1 vs our method.}
    \label{fig:ret_acc_and_ans_acc}
\end{figure}

\subsection{Effect on Retrieval Accuracy}
Figure~\ref{fig:ret_acc} compares the retrieval accuracy of our method and Search-R1 on the test dataset, using checkpoints of the trained models created at different numbers of training steps.
We define retrieval accuracy to be the frequency with which the retrieved documents contain at least one of the ground truth answers.
Specifically, we check whether any of the ground truth answers are present in any if the the retrieved documents via string matching.
We observe that our method consistently achieves higher retrieval accuracy than Search-R1.
This demonstrates the effectiveness of our method in generating better subqueries, which enables the model to retrieve more relevant and accurate information.
The ability to retrieve relevant documents is critical for synthesizing correct answers, particularly in complex, multi-hop question-answering tasks.

\subsection{Effect on Answer Synthesis}
Figure~\ref{fig:ans_acc} compares the accuracy of the final answer, given correct retrieval, for the two methods.
This metric calculates the fraction of correct retrievals that result in a correct answer.
It measures the ability of the model to reason over the retrieved information to generate the correct answer when the retrieved documents contain the ground truth answer.
We observe that our method consistently outperforms Search-R1 with respect to this metric.
This demonstrates our method's ability to train the model to better synthesize the final answer after retrieving the correct documents.
The plots in Figure~\ref{fig:ret_acc_and_ans_acc} demonstrate that our method not only improves the model's capability to generate better subqueries, which lead to better retrievals, but also enhances its ability to combine the retrieved information into a correct final answer.

\subsection{Effect on RL Reward}
\begin{figure}[t!]
    \centering
    \begin{subfigure}[t]{0.23\textwidth}
        \centering
        \includegraphics[width=\textwidth]{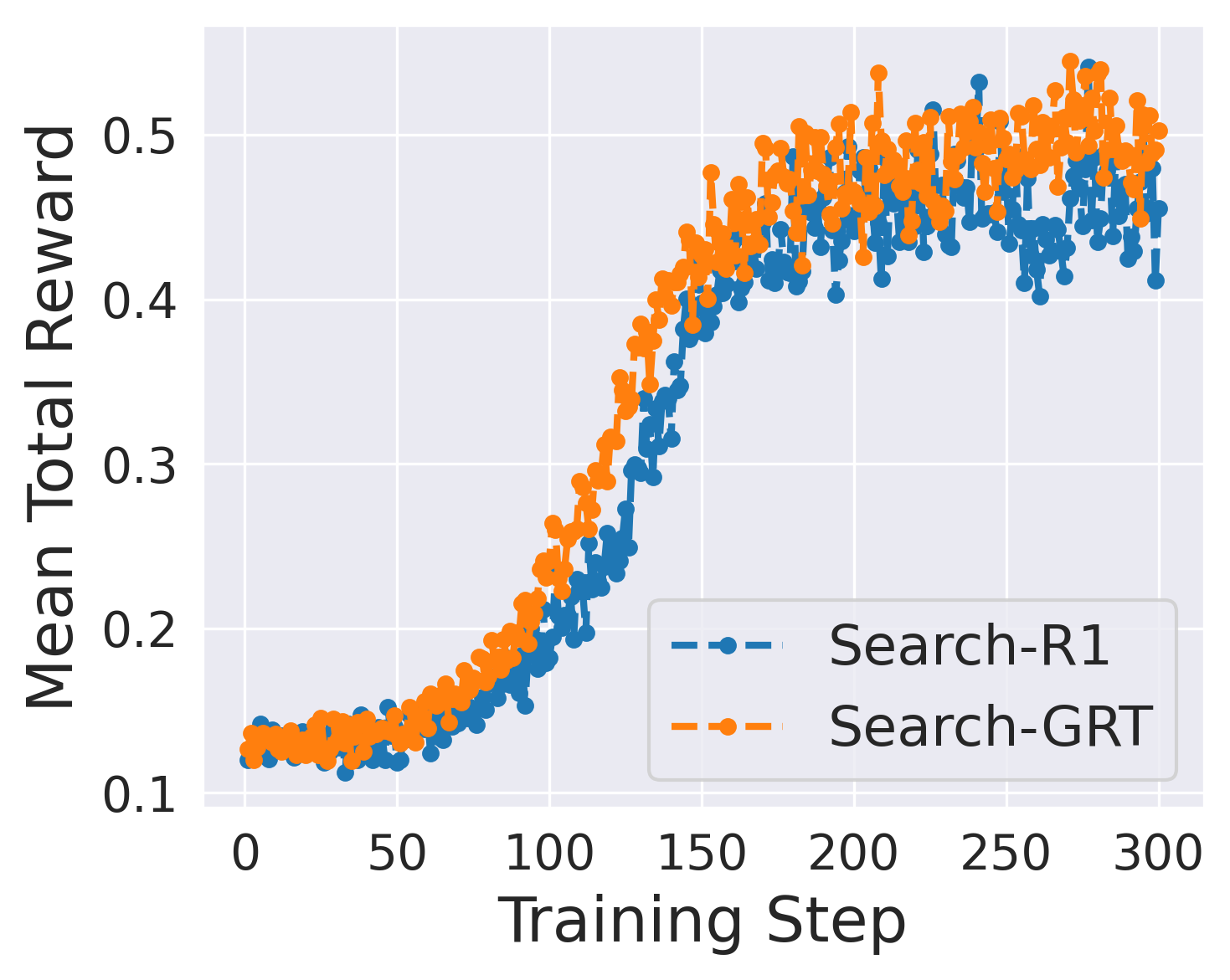}
        \caption{Training Reward}
        \label{fig:training_reward}
    \end{subfigure}%
    ~ 
    \begin{subfigure}[t]{0.23\textwidth}
        \centering
        \includegraphics[width=\textwidth]{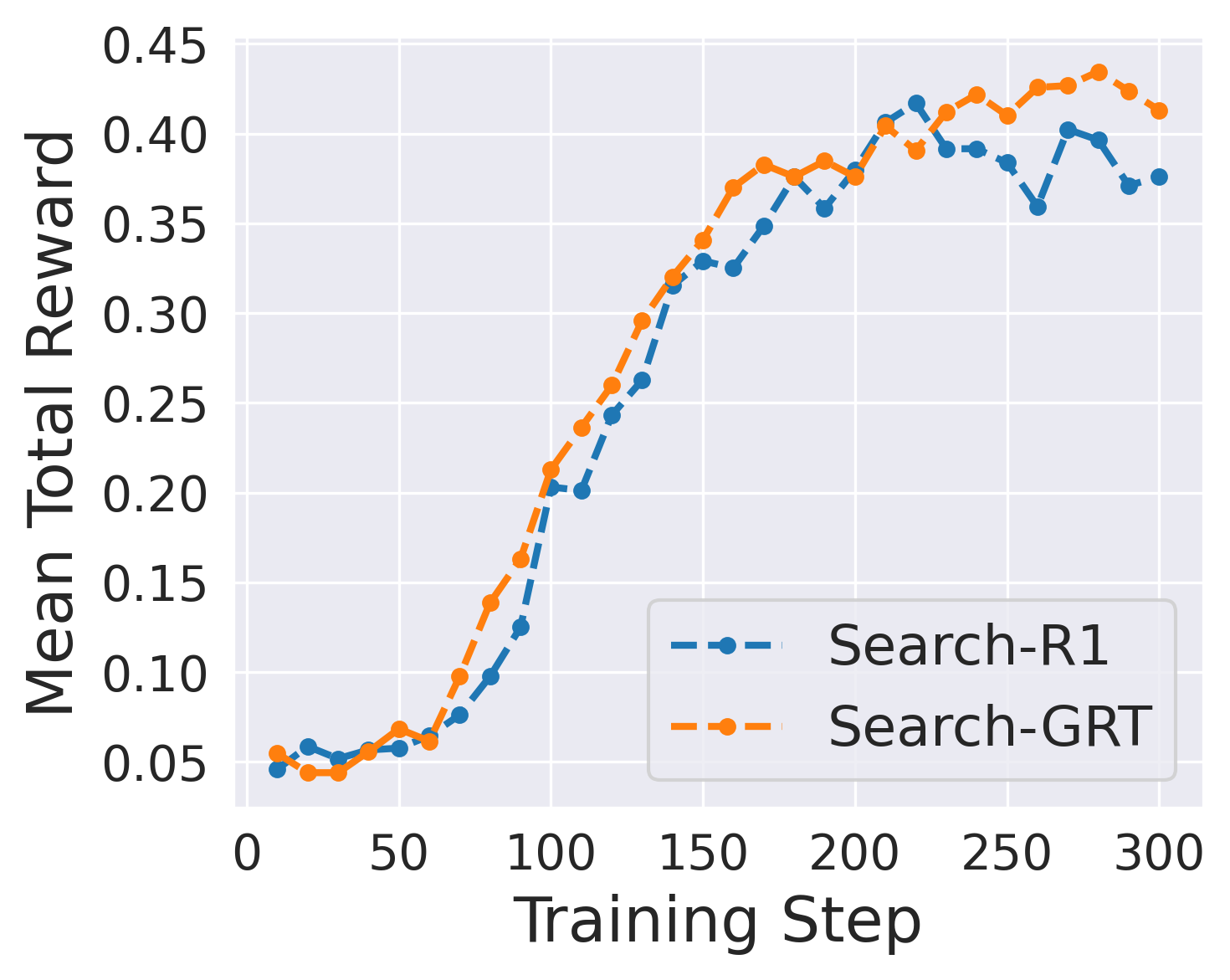}
        \caption{Validation Reward}
        \label{fig:validation_reward}
    \end{subfigure}
    \caption{Comparing RL reward obtained by Search-R1 vs our method on training and validation samples.}
    \label{fig:rl_reward}
\end{figure}

Figure~\ref{fig:rl_reward} compares the RL reward obtained by our method (GRT) and Search-R1 on both training and validation samples. In Figure~\ref{fig:training_reward}, we observe that our GRT approach provides the model with a higher training reward, offering a stronger learning signal to the LLM.
This improvement partially addresses the challenge of sparse rewards in the early stages of the RL training (steps 0-150), enabling the model to learn more effectively.
This enhanced learning signal is achieved through our guided retrieval mechanism, which ensures that the model receives more relevant and accurate information during training, thereby improving its ability to generate correct answers.

At the same time, Figure~\ref{fig:validation_reward} demonstrates that our method achieves higher reward values on validation samples, indicating its ability to generalize to unseen data. Notably, during validation, the retriever operates on the entire Wikipedia 2018 corpus, without the restrictions applied during training.
This setup simulates the actual inference environment, where the model must retrieve information from a vast and unrestricted knowledge base.
This result highlights our method's robustness and effectiveness in utilizing the actual search engine. It demonstrates that our approach does not merely rely on the restricted corpus used during training but instead enhances the LLM's ability to retrieve relevant information from a much larger and more complex knowledge base.

\section{Limitations}
While our Guided Retrieval Training method demonstrates significant improvements in search agent performance, particularly for multi-hop question-answering tasks, it is important to acknowledge its limitations. First, our approach relies on ground truth information to restrict the retrieval corpus during training, which may not always be available for all datasets or real-world applications. This dependency could limit the generalizability of GRT to scenarios where ground truth data is scarce or unavailable. Second, the method is evaluated on a specific set of QA datasets, and its performance on other types of tasks or domains remains unexplored. Future work could investigate the adaptability of GRT to diverse datasets and domains. 

\section{Conclusions}
We introduce Guided Retrieval Training (GRT), a novel method to enhance search agents by improving retrieval and answer synthesis during training.
Our experimental results demonstrate that GRT achieves consistent performance improvements over Search-R1 across a wide range of question-answering (QA) tasks, including both general QA and multi-hop QA (MHQA). For MHQA tasks, our method achieves a greater improvement in performance, highlighting its effectiveness in addressing the unique challenges of complex, multi-hop reasoning.
GRT enhances retrieval accuracy by generating better subqueries and improves answer synthesis, leading to more accurate answers. It provides a stronger learning signal during training and generalizes well to unseen samples, demonstrating its ability to use the actual search engine effectively at inference time.

Beyond search agent research, GRT contributes to AI systems' ability to retrieve and synthesize information from vast knowledge bases, with potential
applications in education, healthcare, and customer support. 
In summary, GRT is a significant step forward in search agent research, offering a robust and efficient solution for complex QA tasks.

\section{Generative AI Use Disclosure}
In the preparation of this paper, generative AI tools like large language models have been used to assist in the editing and polishing of the text. These tools have been employed to enhance the readability, clarity, and language quality of the manuscript, without altering the core research findings, methodologies, or conclusions.
All content presented in this paper has been reviewed and revised by the authors to ensure accuracy and alignment with research objectives.

\bibliographystyle{IEEEtran}
\bibliography{mybib,mybib2}

@article{Gao2023RAGsurvey,
  author       = {Yunfan Gao and
                  Yun Xiong and
                  Xinyu Gao and
                  Kangxiang Jia and
                  Jinliu Pan and
                  Yuxi Bi and
                  Yi Dai and
                  Jiawei Sun and
                  Qianyu Guo and
                  Meng Wang and
                  Haofen Wang},
  title        = {Retrieval-Augmented Generation for Large Language Models: {A} Survey},
  journal      = {CoRR},
  volume       = {abs/2312.10997},
  year         = {2023},
  url          = {https://doi.org/10.48550/arXiv.2312.10997},
  doi          = {10.48550/ARXIV.2312.10997},
  eprinttype    = {arXiv},
  eprint       = {2312.10997},
  bibsource    = {dblp computer science bibliography, https://dblp.org}
}

@article{search_arena_2025,
  author       = {Mihran Miroyan and
                  Tsung{-}Han Wu and
                  Logan King and
                  Tianle Li and
                  Jiayi Pan and
                  Xinyan Hu and
                  Wei{-}Lin Chiang and
                  Anastasios N. Angelopoulos and
                  Trevor Darrell and
                  Narges Norouzi and
                  Joseph E. Gonzalez},
  title        = {Search Arena: Analyzing Search-Augmented LLMs},
  journal      = {CoRR},
  volume       = {abs/2506.05334},
  year         = {2025},
  url          = {https://doi.org/10.48550/arXiv.2506.05334},
  doi          = {10.48550/ARXIV.2506.05334},
  eprinttype    = {arXiv},
  eprint       = {2506.05334},
  bibsource    = {dblp computer science bibliography, https://dblp.org}
}

@inproceedings{Fan2024RAGSurvey,
author = {Fan, Wenqi and Ding, Yujuan and Ning, Liangbo and Wang, Shijie and Li, Hengyun and Yin, Dawei and Chua, Tat-Seng and Li, Qing},
title = {A Survey on RAG Meeting LLMs: Towards Retrieval-Augmented Large Language Models},
year = {2024},
isbn = {9798400704901},
publisher = {Association for Computing Machinery},
address = {New York, NY, USA},
url = {https://doi.org/10.1145/3637528.3671470},
doi = {10.1145/3637528.3671470},
booktitle = {Proceedings of the 30th ACM SIGKDD Conference on Knowledge Discovery and Data Mining},
pages = {6491–6501},
numpages = {11},
location = {Barcelona, Spain},
series = {KDD '24}
}

@misc{liu2025genrewrite,
      title={GenRewrite: Query Rewriting via Large Language Models}, 
      author={Jie Liu and Barzan Mozafari},
      year={2025},
      eprint={2403.09060},
      archivePrefix={arXiv},
      primaryClass={cs.DB},
      url={https://arxiv.org/abs/2403.09060}, 
}

@inproceedings{ye-etal-2023-enhancing,
    title = "Enhancing Conversational Search: Large Language Model-Aided Informative Query Rewriting",
    author = "Ye, Fanghua  and
      Fang, Meng  and
      Li, Shenghui  and
      Yilmaz, Emine",
    editor = "Bouamor, Houda  and
      Pino, Juan  and
      Bali, Kalika",
    booktitle = "Findings of the Association for Computational Linguistics: EMNLP 2023",
    month = dec,
    year = "2023",
    address = "Singapore",
    publisher = "Association for Computational Linguistics",
    url = "https://aclanthology.org/2023.findings-emnlp.398/",
    doi = "10.18653/v1/2023.findings-emnlp.398",
    pages = "5985--6006"
}

@inproceedings{ma-etal-2023-query,
    title = "Query Rewriting in Retrieval-Augmented Large Language Models",
    author = "Ma, Xinbei  and
      Gong, Yeyun  and
      He, Pengcheng  and
      Zhao, Hai  and
      Duan, Nan",
    editor = "Bouamor, Houda  and
      Pino, Juan  and
      Bali, Kalika",
    booktitle = "Proceedings of the 2023 Conference on Empirical Methods in Natural Language Processing",
    month = dec,
    year = "2023",
    address = "Singapore",
    publisher = "Association for Computational Linguistics",
    url = "https://aclanthology.org/2023.emnlp-main.322/",
    doi = "10.18653/v1/2023.emnlp-main.322",
    pages = "5303--5315"
}

@inproceedings{MaGHZD23,
  author       = {Xinbei Ma and
                  Yeyun Gong and
                  Pengcheng He and
                  Hai Zhao and
                  Nan Duan},
  editor       = {Houda Bouamor and
                  Juan Pino and
                  Kalika Bali},
  title        = {Query Rewriting in Retrieval-Augmented Large Language Models},
  booktitle    = {Proceedings of the 2023 Conference on Empirical Methods in Natural
                  Language Processing, {EMNLP} 2023, Singapore, December 6-10, 2023},
  pages        = {5303--5315},
  publisher    = {Association for Computational Linguistics},
  year         = {2023},
  url          = {https://doi.org/10.18653/v1/2023.emnlp-main.322},
  doi          = {10.18653/V1/2023.EMNLP-MAIN.322},
  bibsource    = {dblp computer science bibliography, https://dblp.org}
}

@article{Wang-e5-2022,
  author       = {Liang Wang and
                  Nan Yang and
                  Xiaolong Huang and
                  Binxing Jiao and
                  Linjun Yang and
                  Daxin Jiang and
                  Rangan Majumder and
                  Furu Wei},
  title        = {Text Embeddings by Weakly-Supervised Contrastive Pre-training},
  journal      = {CoRR},
  volume       = {abs/2212.03533},
  year         = {2022},
  url          = {https://doi.org/10.48550/arXiv.2212.03533},
  doi          = {10.48550/ARXIV.2212.03533},
  eprinttype    = {arXiv},
  eprint       = {2212.03533},
  bibsource    = {dblp computer science bibliography, https://dblp.org}
}

@inproceedings{karpukhin-etal-2020-dense,
    title = "Dense Passage Retrieval for Open-Domain Question Answering",
    author = "Karpukhin, Vladimir  and
      Oguz, Barlas  and
      Min, Sewon  and
      Lewis, Patrick  and
      Wu, Ledell  and
      Edunov, Sergey  and
      Chen, Danqi  and
      Yih, Wen-tau",
    editor = "Webber, Bonnie  and
      Cohn, Trevor  and
      He, Yulan  and
      Liu, Yang",
    booktitle = "Proceedings of the 2020 Conference on Empirical Methods in Natural Language Processing (EMNLP)",
    month = nov,
    year = "2020",
    address = "Online",
    publisher = "Association for Computational Linguistics",
    url = "https://aclanthology.org/2020.emnlp-main.550/",
    doi = "10.18653/v1/2020.emnlp-main.550",
    pages = "6769--6781"
}

@article{ppo-SchulmanWDRK17,
  author       = {John Schulman and
                  Filip Wolski and
                  Prafulla Dhariwal and
                  Alec Radford and
                  Oleg Klimov},
  title        = {Proximal Policy Optimization Algorithms},
  journal      = {CoRR},
  volume       = {abs/1707.06347},
  year         = {2017},
  url          = {http://arxiv.org/abs/1707.06347},
  eprinttype    = {arXiv},
  eprint       = {1707.06347},
  bibsource    = {dblp computer science bibliography, https://dblp.org}
}

@article{Yang-qwen-2024,
  author       = {An Yang and
                  Baosong Yang and
                  Beichen Zhang and
                  Binyuan Hui and
                  Bo Zheng and
                  Bowen Yu and
                  Chengyuan Li and
                  Dayiheng Liu and
                  Fei Huang and
                  Haoran Wei and
                  Huan Lin and
                  Jian Yang and
                  Jianhong Tu and
                  Jianwei Zhang and
                  Jianxin Yang and
                  Jiaxi Yang and
                  Jingren Zhou and
                  Junyang Lin and
                  Kai Dang and
                  Keming Lu and
                  Keqin Bao and
                  Kexin Yang and
                  Le Yu and
                  Mei Li and
                  Mingfeng Xue and
                  Pei Zhang and
                  Qin Zhu and
                  Rui Men and
                  Runji Lin and
                  Tianhao Li and
                  Tingyu Xia and
                  Xingzhang Ren and
                  Xuancheng Ren and
                  Yang Fan and
                  Yang Su and
                  Yichang Zhang and
                  Yu Wan and
                  Yuqiong Liu and
                  Zeyu Cui and
                  Zhenru Zhang and
                  Zihan Qiu},
  title        = {Qwen2.5 Technical Report},
  journal      = {CoRR},
  volume       = {abs/2412.15115},
  year         = {2024},
  url          = {https://doi.org/10.48550/arXiv.2412.15115},
  doi          = {10.48550/ARXIV.2412.15115},
  eprinttype    = {arXiv},
  eprint       = {2412.15115},
  bibsource    = {dblp computer science bibliography, https://dblp.org}
}

@article{yuan-rft-2023,
  author       = {Zheng Yuan and
                  Hongyi Yuan and
                  Chengpeng Li and
                  Guanting Dong and
                  Chuanqi Tan and
                  Chang Zhou},
  title        = {Scaling Relationship on Learning Mathematical Reasoning with Large
                  Language Models},
  journal      = {CoRR},
  volume       = {abs/2308.01825},
  year         = {2023},
  url          = {https://doi.org/10.48550/arXiv.2308.01825},
  doi          = {10.48550/ARXIV.2308.01825},
  eprinttype    = {arXiv},
  eprint       = {2308.01825},
  bibsource    = {dblp computer science bibliography, https://dblp.org}
}

@article{GuoYZSWZXZMBZY025,
  author       = {Daya Guo and
                  Dejian Yang and
                  Haowei Zhang and
                  Junxiao Song and
                  Peiyi Wang and
                  Qihao Zhu and
                  Runxin Xu and
                  Ruoyu Zhang and
                  Shirong Ma and
                  Xiao Bi and
                  Xiaokang Zhang and
                  Xingkai Yu and
                  Yu Wu and
                  Z. F. Wu and
                  Zhibin Gou and
                  Zhihong Shao and
                  Zhuoshu Li and
                  Ziyi Gao and
                  Aixin Liu and
                  Bing Xue and
                  Bingxuan Wang and
                  Bochao Wu and
                  Bei Feng and
                  Chengda Lu and
                  Chenggang Zhao and
                  Chengqi Deng and
                  Chong Ruan and
                  Damai Dai and
                  Deli Chen and
                  Dongjie Ji and
                  Erhang Li and
                  Fangyun Lin and
                  Fucong Dai and
                  Fuli Luo and
                  Guangbo Hao and
                  Guanting Chen and
                  Guowei Li and
                  Hao Zhang and
                  Hanwei Xu and
                  Honghui Ding and
                  Huazuo Gao and
                  Hui Qu and
                  Hui Li and
                  Jianzhong Guo and
                  Jiashi Li and
                  Jingchang Chen and
                  Jingyang Yuan and
                  Jinhao Tu and
                  Junjie Qiu and
                  Junlong Li and
                  J. L. Cai and
                  Jiaqi Ni and
                  Jian Liang and
                  Jin Chen and
                  Kai Dong and
                  Kai Hu and
                  Kaichao You and
                  Kaige Gao and
                  Kang Guan and
                  Kexin Huang and
                  Kuai Yu and
                  Lean Wang and
                  Lecong Zhang and
                  Liang Zhao and
                  Litong Wang and
                  Liyue Zhang and
                  Lei Xu and
                  Leyi Xia and
                  Mingchuan Zhang and
                  Minghua Zhang and
                  Minghui Tang and
                  Mingxu Zhou and
                  Meng Li and
                  Miaojun Wang and
                  Mingming Li and
                  Ning Tian and
                  Panpan Huang and
                  Peng Zhang and
                  Qiancheng Wang and
                  Qinyu Chen and
                  Qiushi Du and
                  Ruiqi Ge and
                  Ruisong Zhang and
                  Ruizhe Pan and
                  Runji Wang and
                  R. J. Chen and
                  R. L. Jin and
                  Ruyi Chen and
                  Shanghao Lu and
                  Shangyan Zhou and
                  Shanhuang Chen and
                  Shengfeng Ye and
                  Shiyu Wang and
                  Shuiping Yu and
                  Shunfeng Zhou and
                  Shuting Pan and
                  S. S. Li and
                  Shuang Zhou and
                  Shaoqing Wu and
                  Tao Yun and
                  Tian Pei and
                  Tianyu Sun and
                  Tao Wang and
                  Wangding Zeng and
                  Wen Liu and
                  Wenfeng Liang and
                  Wenjun Gao and
                  Wenqin Yu and
                  Wentao Zhang and
                  W. L. Xiao and
                  Wei An and
                  Xiaodong Liu and
                  Xiaohan Wang and
                  Xiaokang Chen and
                  Xiaotao Nie and
                  Xin Cheng and
                  Xin Liu and
                  Xin Xie and
                  Xingchao Liu and
                  Xinyu Yang and
                  Xinyuan Li and
                  Xuecheng Su and
                  Xuheng Lin and
                  X. Q. Li and
                  Xiangyue Jin and
                  Xiaojin Shen and
                  Xiaosha Chen and
                  Xiaowen Sun and
                  Xiaoxiang Wang and
                  Xinnan Song and
                  Xinyi Zhou and
                  Xianzu Wang and
                  Xinxia Shan and
                  Y. K. Li and
                  Y. Q. Wang and
                  Y. X. Wei and
                  Yang Zhang and
                  Yanhong Xu and
                  Yao Li and
                  Yao Zhao and
                  Yaofeng Sun and
                  Yaohui Wang and
                  Yi Yu and
                  Yichao Zhang and
                  Yifan Shi and
                  Yiliang Xiong and
                  Ying He and
                  Yishi Piao and
                  Yisong Wang and
                  Yixuan Tan and
                  Yiyang Ma and
                  Yiyuan Liu and
                  Yongqiang Guo and
                  Yuan Ou and
                  Yuduan Wang and
                  Yue Gong and
                  Yuheng Zou and
                  Yujia He and
                  Yunfan Xiong and
                  Yuxiang Luo and
                  Yuxiang You and
                  Yuxuan Liu and
                  Yuyang Zhou and
                  Y. X. Zhu and
                  Yanping Huang and
                  Yaohui Li and
                  Yi Zheng and
                  Yuchen Zhu and
                  Yunxian Ma and
                  Ying Tang and
                  Yukun Zha and
                  Yuting Yan and
                  Z. Z. Ren and
                  Zehui Ren and
                  Zhangli Sha and
                  Zhe Fu and
                  Zhean Xu and
                  Zhenda Xie and
                  Zhengyan Zhang and
                  Zhewen Hao and
                  Zhicheng Ma and
                  Zhigang Yan and
                  Zhiyu Wu and
                  Zihui Gu and
                  Zijia Zhu and
                  Zijun Liu and
                  Zilin Li and
                  Ziwei Xie and
                  Ziyang Song and
                  Zizheng Pan and
                  Zhen Huang and
                  Zhipeng Xu and
                  Zhongyu Zhang and
                  Zhen Zhang},
  title        = {DeepSeek-R1 incentivizes reasoning in LLMs through reinforcement learning},
  journal      = {Nat.},
  volume       = {645},
  number       = {8081},
  pages        = {633--638},
  year         = {2025},
  url          = {https://doi.org/10.1038/s41586-025-09422-z},
  doi          = {10.1038/S41586-025-09422-Z},
  bibsource    = {dblp computer science bibliography, https://dblp.org}
}

@article{ChungHLZTFL00BW24,
  author       = {Hyung Won Chung and
                  Le Hou and
                  Shayne Longpre and
                  Barret Zoph and
                  Yi Tay and
                  William Fedus and
                  Yunxuan Li and
                  Xuezhi Wang and
                  Mostafa Dehghani and
                  Siddhartha Brahma and
                  Albert Webson and
                  Shixiang Shane Gu and
                  Zhuyun Dai and
                  Mirac Suzgun and
                  Xinyun Chen and
                  Aakanksha Chowdhery and
                  Alex Castro{-}Ros and
                  Marie Pellat and
                  Kevin Robinson and
                  Dasha Valter and
                  Sharan Narang and
                  Gaurav Mishra and
                  Adams Yu and
                  Vincent Y. Zhao and
                  Yanping Huang and
                  Andrew M. Dai and
                  Hongkun Yu and
                  Slav Petrov and
                  Ed H. Chi and
                  Jeff Dean and
                  Jacob Devlin and
                  Adam Roberts and
                  Denny Zhou and
                  Quoc V. Le and
                  Jason Wei},
  title        = {Scaling Instruction-Finetuned Language Models},
  journal      = {J. Mach. Learn. Res.},
  volume       = {25},
  pages        = {70:1--70:53},
  year         = {2024},
  url          = {https://jmlr.org/papers/v25/23-0870.html},
  bibsource    = {dblp computer science bibliography, https://dblp.org}
}

@inproceedings{Li-search-o1-2025,
  author       = {Xiaoxi Li and
                  Guanting Dong and
                  Jiajie Jin and
                  Yuyao Zhang and
                  Yujia Zhou and
                  Yutao Zhu and
                  Peitian Zhang and
                  Zhicheng Dou},
  editor       = {Christos Christodoulopoulos and
                  Tanmoy Chakraborty and
                  Carolyn Rose and
                  Violet Peng},
  title        = {Search-o1: Agentic Search-Enhanced Large Reasoning Models},
  booktitle    = {Proceedings of the 2025 Conference on Empirical Methods in Natural
                  Language Processing, {EMNLP} 2025, Suzhou, China, November 4-9, 2025},
  pages        = {5420--5438},
  publisher    = {Association for Computational Linguistics},
  year         = {2025},
  url          = {https://doi.org/10.18653/v1/2025.emnlp-main.276},
  doi          = {10.18653/V1/2025.EMNLP-MAIN.276},
  bibsource    = {dblp computer science bibliography, https://dblp.org}
}

@inproceedings{TrivediBKS23,
  author       = {Harsh Trivedi and
                  Niranjan Balasubramanian and
                  Tushar Khot and
                  Ashish Sabharwal},
  editor       = {Anna Rogers and
                  Jordan L. Boyd{-}Graber and
                  Naoaki Okazaki},
  title        = {Interleaving Retrieval with Chain-of-Thought Reasoning for Knowledge-Intensive
                  Multi-Step Questions},
  booktitle    = {Proceedings of the 61st Annual Meeting of the Association for Computational
                  Linguistics (Volume 1: Long Papers), {ACL} 2023, Toronto, Canada,
                  July 9-14, 2023},
  pages        = {10014--10037},
  publisher    = {Association for Computational Linguistics},
  year         = {2023},
  url          = {https://doi.org/10.18653/v1/2023.acl-long.557},
  doi          = {10.18653/V1/2023.ACL-LONG.557},
  bibsource    = {dblp computer science bibliography, https://dblp.org}
}

@inproceedings{rag_lewis_2020,
author = {Lewis, Patrick and Perez, Ethan and Piktus, Aleksandra and Petroni, Fabio and Karpukhin, Vladimir and Goyal, Naman and K\"{u}ttler, Heinrich and Lewis, Mike and Yih, Wen-tau and Rockt\"{a}schel, Tim and Riedel, Sebastian and Kiela, Douwe},
title = {Retrieval-augmented generation for knowledge-intensive NLP tasks},
year = {2020},
isbn = {9781713829546},
publisher = {Curran Associates Inc.},
address = {Red Hook, NY, USA},
booktitle = {Proceedings of the 34th International Conference on Neural Information Processing Systems},
articleno = {793},
numpages = {16},
location = {Vancouver, BC, Canada},
series = {NIPS '20}
}

@inproceedings{cot_wei_2022,
author = {Wei, Jason and Wang, Xuezhi and Schuurmans, Dale and Bosma, Maarten and Ichter, Brian and Xia, Fei and Chi, Ed H. and Le, Quoc V. and Zhou, Denny},
title = {Chain-of-thought prompting elicits reasoning in large language models},
year = {2022},
isbn = {9781713871088},
publisher = {Curran Associates Inc.},
address = {Red Hook, NY, USA},
booktitle = {Proceedings of the 36th International Conference on Neural Information Processing Systems},
articleno = {1800},
numpages = {14},
location = {New Orleans, LA, USA},
series = {NIPS '22}
}

@inproceedings{press-etal-2023-measuring,
    title = "Measuring and Narrowing the Compositionality Gap in Language Models",
    author = "Press, Ofir  and
      Zhang, Muru  and
      Min, Sewon  and
      Schmidt, Ludwig  and
      Smith, Noah  and
      Lewis, Mike",
    editor = "Bouamor, Houda  and
      Pino, Juan  and
      Bali, Kalika",
    booktitle = "Findings of the Association for Computational Linguistics: EMNLP 2023",
    month = dec,
    year = "2023",
    address = "Singapore",
    publisher = "Association for Computational Linguistics",
    url = "https://aclanthology.org/2023.findings-emnlp.378/",
    doi = "10.18653/v1/2023.findings-emnlp.378",
    pages = "5687--5711"
}

@article{trivedi-etal-2022-musique,
    title = "{M}u{S}i{Q}ue: Multihop Questions via Single-hop Question Composition",
    author = "Trivedi, Harsh  and
      Balasubramanian, Niranjan  and
      Khot, Tushar  and
      Sabharwal, Ashish",
    editor = "Roark, Brian  and
      Nenkova, Ani",
    journal = "Transactions of the Association for Computational Linguistics",
    volume = "10",
    year = "2022",
    address = "Cambridge, MA",
    publisher = "MIT Press",
    url = "https://aclanthology.org/2022.tacl-1.31/",
    doi = "10.1162/tacl_a_00475",
    pages = "539--554"
}

@inproceedings{ho-etal-2020-constructing,
    title = "Constructing A Multi-hop {QA} Dataset for Comprehensive Evaluation of Reasoning Steps",
    author = "Ho, Xanh  and
      Duong Nguyen, Anh-Khoa  and
      Sugawara, Saku  and
      Aizawa, Akiko",
    editor = "Scott, Donia  and
      Bel, Nuria  and
      Zong, Chengqing",
    booktitle = "Proceedings of the 28th International Conference on Computational Linguistics",
    month = dec,
    year = "2020",
    address = "Barcelona, Spain (Online)",
    publisher = "International Committee on Computational Linguistics",
    url = "https://aclanthology.org/2020.coling-main.580/",
    doi = "10.18653/v1/2020.coling-main.580",
    pages = "6609--6625"
}

@inproceedings{yang-etal-2018-hotpotqa,
    title = "{H}otpot{QA}: A Dataset for Diverse, Explainable Multi-hop Question Answering",
    author = "Yang, Zhilin  and
      Qi, Peng  and
      Zhang, Saizheng  and
      Bengio, Yoshua  and
      Cohen, William  and
      Salakhutdinov, Ruslan  and
      Manning, Christopher D.",
    editor = "Riloff, Ellen  and
      Chiang, David  and
      Hockenmaier, Julia  and
      Tsujii, Jun{'}ichi",
    booktitle = "Proceedings of the 2018 Conference on Empirical Methods in Natural Language Processing",
    month = oct # "-" # nov,
    year = "2018",
    address = "Brussels, Belgium",
    publisher = "Association for Computational Linguistics",
    url = "https://aclanthology.org/D18-1259/",
    doi = "10.18653/v1/D18-1259",
    pages = "2369--2380"
}

@inproceedings{mallen-etal-2023-trust,
    title = "When Not to Trust Language Models: Investigating Effectiveness of Parametric and Non-Parametric Memories",
    author = "Mallen, Alex  and
      Asai, Akari  and
      Zhong, Victor  and
      Das, Rajarshi  and
      Khashabi, Daniel  and
      Hajishirzi, Hannaneh",
    editor = "Rogers, Anna  and
      Boyd-Graber, Jordan  and
      Okazaki, Naoaki",
    booktitle = "Proceedings of the 61st Annual Meeting of the Association for Computational Linguistics (Volume 1: Long Papers)",
    month = jul,
    year = "2023",
    address = "Toronto, Canada",
    publisher = "Association for Computational Linguistics",
    url = "https://aclanthology.org/2023.acl-long.546/",
    doi = "10.18653/v1/2023.acl-long.546",
    pages = "9802--9822"
}

@inproceedings{joshi-etal-2017-triviaqa,
    title = "{T}rivia{QA}: A Large Scale Distantly Supervised Challenge Dataset for Reading Comprehension",
    author = "Joshi, Mandar  and
      Choi, Eunsol  and
      Weld, Daniel  and
      Zettlemoyer, Luke",
    editor = "Barzilay, Regina  and
      Kan, Min-Yen",
    booktitle = "Proceedings of the 55th Annual Meeting of the Association for Computational Linguistics (Volume 1: Long Papers)",
    month = jul,
    year = "2017",
    address = "Vancouver, Canada",
    publisher = "Association for Computational Linguistics",
    url = "https://aclanthology.org/P17-1147/",
    doi = "10.18653/v1/P17-1147",
    pages = "1601--1611"
}

@article{kwiatkowski-etal-2019-natural,
    title = "Natural Questions: A Benchmark for Question Answering Research",
    author = "Kwiatkowski, Tom  and
      Palomaki, Jennimaria  and
      Redfield, Olivia  and
      Collins, Michael  and
      Parikh, Ankur  and
      Alberti, Chris  and
      Epstein, Danielle  and
      Polosukhin, Illia  and
      Devlin, Jacob  and
      Lee, Kenton  and
      Toutanova, Kristina  and
      Jones, Llion  and
      Kelcey, Matthew  and
      Chang, Ming-Wei  and
      Dai, Andrew M.  and
      Uszkoreit, Jakob  and
      Le, Quoc  and
      Petrov, Slav",
    editor = "Lee, Lillian  and
      Johnson, Mark  and
      Roark, Brian  and
      Nenkova, Ani",
    journal = "Transactions of the Association for Computational Linguistics",
    volume = "7",
    year = "2019",
    address = "Cambridge, MA",
    publisher = "MIT Press",
    url = "https://aclanthology.org/Q19-1026/",
    doi = "10.1162/tacl_a_00276",
    pages = "452--466"
}

@inproceedings{
jin2025searchr,
title={Search-R1: Training {LLM}s to Reason and Leverage Search Engines with Reinforcement Learning},
author={Bowen Jin and Hansi Zeng and Zhenrui Yue and Jinsung Yoon and Sercan O Arik and Dong Wang and Hamed Zamani and Jiawei Han},
booktitle={Second Conference on Language Modeling},
year={2025},
url={https://openreview.net/forum?id=Rwhi91ideu}
}

@article{sun2025zerosearch,
  title={Zerosearch: Incentivize the search capability of llms without searching},
  author={Sun, Hao and Qiao, Zile and Guo, Jiayan and Fan, Xuanbo and Hou, Yingyan and Jiang, Yong and Xie, Pengjun and Zhang, Yan and Huang, Fei and Zhou, Jingren},
  journal={arXiv preprint arXiv:2505.04588},
  year={2025}
}

@inproceedings{chen2017reading,
  title={Reading wikipedia to answer open-domain questions},
  author={Chen, Danqi and Fisch, Adam and Weston, Jason and Bordes, Antoine},
  booktitle={Proceedings of the 55th Annual Meeting of the Association for Computational Linguistics (Volume 1: Long Papers)},
  pages={1870--1879},
  year={2017}
}

@inproceedings{yang2018hotpotqa,
  title={HotpotQA: A dataset for diverse, explainable multi-hop question answering},
  author={Yang, Zhilin and Qi, Peng and Zhang, Saizheng and Bengio, Yoshua and Cohen, William and Salakhutdinov, Ruslan and Manning, Christopher D},
  booktitle={Proceedings of the 2018 conference on empirical methods in natural language processing},
  pages={2369--2380},
  year={2018}
}

@inproceedings{trivedi2023interleaving,
  title={Interleaving retrieval with chain-of-thought reasoning for knowledge-intensive multi-step questions},
  author={Trivedi, Harsh and Balasubramanian, Niranjan and Khot, Tushar and Sabharwal, Ashish},
  booktitle={Proceedings of the 61st annual meeting of the association for computational linguistics (volume 1: long papers)},
  pages={10014--10037},
  year={2023}
}

@inproceedings{yao2022react,
  title={React: Synergizing reasoning and acting in language models},
  author={Yao, Shunyu and Zhao, Jeffrey and Yu, Dian and Du, Nan and Shafran, Izhak and Narasimhan, Karthik R and Cao, Yuan},
  booktitle={The eleventh international conference on learning representations},
  year={2022}
}

@article{wang2025chain,
  title={Chain-of-Retrieval Augmented Generation},
  author={Wang, Liang and Chen, Haonan and Yang, Nan and Huang, Xiaolong and Dou, Zhicheng and Wei, Furu},
  journal={arXiv preprint arXiv:2501.14342},
  year={2025}
}

\end{document}